%% file: neurips_2025.tex
\documentclass{article}

\usepackage[nonatbib,dblblindworkshop,final]{neurips_2025}

\usepackage[utf8]{inputenc} 
\usepackage[T1]{fontenc}    
\usepackage{hyperref}       
\usepackage{url}            
\usepackage{booktabs}       
\usepackage{amsfonts}       
\usepackage{nicefrac}       
\usepackage{microtype}      
\usepackage{xcolor}         
\usepackage{amsmath}
\usepackage{amssymb}
\usepackage{algorithm}
\usepackage{algpseudocode} 
\usepackage{listings}
\usepackage{tabularx}
\usepackage{adjustbox}
\usepackage{graphicx}
\usepackage{wrapfig}
\usepackage{enumitem}
\usepackage{subcaption} 
\usepackage{array}
\setlist[itemize]{leftmargin=*, labelsep=0.4em, nosep}

\usepackage[
  backend=biber,      
  style=numeric,      
  sorting=none,
  natbib=true         
]{biblatex}
\let\cite\parencite

\workshoptitle{Evaluating the Evolving LLM Lifecycle - Benchmarks, Emergent Abilities, and Scaling}
\title{VLM-SlideEval: Evaluating VLMs on Structured Comprehension and Perturbation Sensitivity in PPT}

\author{%
  Hyeonsu B. Kang \qquad Emily Bao \qquad Anjan Goswami\\
  PowerPoint AI, Microsoft Inc., San Francisco, CA 94103\\
  \texttt{\{hyeonsukang, baoemily, anjangoswami\}@microsoft.com}
}

\input{src/macros}

\begin{document}

\maketitle

\input{src/00_abstract}

\input{src/01_intro}
\input{src/02_related_work}

\input{src/03_method}

\input{src/04_results}

\input{src/05_conclusion}

\printbibliography

\newpage
\appendix
\input{src/06_appendix}

\begin{ack}
We thank Naiqing Guan, Saket Gurukar, Md Muksitul Haque, Jinghua Yao, Lixiang Li, Ruolin Su, Sharena Pari-Monasch, Jesse Harvey, Dan Swett, and the anonymous NeurIPS reviewers for their constructive feedback.
\end{ack}
\end{document}

%% file: src/macros.tex
\newcommand{\eg}{\textit{e.g., }}
\newcommand{\cf}{\textit{cf. }}

\newcommand{\ie}{\textit{i.e., }}

\newcommand{\xhdr}[1]{\vspace{.2em}\noindent{{\bf #1.}}}

\newcommand{\revised}[2]{%
  \ifx\relax#1\relax
    \textcolor{black}{#2}%
  \else
    \textcolor{black}{#2}%
  \fi
}

\definecolor{lightgrey}{HTML}{f0f0f0}

\lstdefinestyle{prompt}{
  basicstyle=\ttfamily\fontsize{8}{11}\selectfont,
  breaklines=true,
  columns=fullflexible,
  basewidth=0.95em,
  linewidth=\textwidth,
  xleftmargin=0pt, xrightmargin=0pt,
  framexleftmargin=0pt, framexrightmargin=0pt,
  framesep=3pt,
  emphstyle=\itshape,
  escapeinside={(*@}{@*)},
}

\lstdefinestyle{code}{
  basicstyle=\ttfamily\fontsize{6}{9}\selectfont,
  breaklines=true,
  frame=single,
  columns=fullflexible,
  basewidth=0.95em,
  aboveskip=1em, belowskip=1em,
  linewidth=\linewidth,
  xleftmargin=0pt, xrightmargin=0pt,
  framexleftmargin=0pt, framexrightmargin=0pt,
  emphstyle=\itshape,
  escapeinside={(*@}{@*)},
}

\newcolumntype{T}{>{\ttfamily\raggedright\arraybackslash}l} 

\newcolumntype{Y}{>{\raggedright\arraybackslash}X} 

\newcommand{\panelheight}{1.2in} 

\renewcommand{\arraystretch}{1.2} 

\urldef{\hfurl}\url{https://huggingface.co/datasets/Forceless/Zenodo10K/viewer/default/pptx}

%% file: src/00_abstract.tex
\begin{abstract}
Vision-language models (VLMs) are increasingly used to evaluate multimodal content, including presentation slides, yet their slide-specific understanding remains underexplored {despite their growing role as critics in agentic, model-forward pipelines}.
We introduce \textbf{VLM-SlideEval}, an evaluation framework that probes VLMs along three axes: (1) element-level extraction from slide images aligned to ground truth; (2) robustness to controlled perturbations in geometry, style, and text; and (3) higher-level comprehension, such as recovering a deck's narrative order from shuffled slides.
Using publicly available decks from Zenodo\footnote{\scriptsize{\url{https://zenodo.org}; HF viewer: \hfurl}}, we standardize ground-truth element metadata from PowerPoint XML and live renderings into a unified, verifiable schema.
Empirically, VLMs underperform on pixel-accurate extraction and show non-trivial agreement, fidelity, and consistency under controlled perturbations, while performing better on single-slide content understanding; however, they do not reliably capture narrative structure across slides.
These results highlight the limits of current VLMs for slide evaluation and motivate calibrated, critic-in-the-loop evaluators that drive iterative refinement and selection in agentic pipelines.
\end{abstract}

%% file: src/01_intro.tex
\vspace{-1.5em}
\section{Introduction} \vspace{-1em}
Presentation slides are a primary vehicle for conveying structured ideas across domains ranging from education to scientific communication to corporate decision-making.
Automatic evaluation of slide quality and content understanding is an emerging and pronounced need, particularly in light of advances in \emph{agentic, model-forward} slide generation~\cite{Ge_2025_CVPR,fu2022doc2ppt}.
While prior work on document analysis has focused on optical character recognition (OCR)~\cite{xu2020layoutlmv2,wang2024docllm,TessOverview} and XML-based parsing~\cite{python_pptx}, these approaches are brittle when slides are only available as rendered images, and are limited to low-level layout information without reasoning about higher-level semantics.
In contrast, vision-language models (VLMs) promise a unified mechanism for parsing slide content directly from images while also supporting tasks that require semantic or narrative comprehension.

Despite the promise, it remains unclear to what extent current VLMs truly comprehend presentation slides.
On one hand, VLMs may struggle with precise pixel-level tasks such as identifying bounding boxes, font attributes, or alignment, since they may not have been directly trained on raw presentation rendering pipelines or large scale OCR data of slide presentations.
On the other hand, VLMs may excel at higher-level understanding, such as identifying the role of slide elements (\eg~title, subtitle, body text), inferring content hierarchy, or reasoning over narrative flow in a deck.
Understanding these trade-offs is crucial for designing reliable and scalable evaluation pipelines that utilize VLMs.

We introduce \textbf{VLM-SlideEval} as a first-class \emph{critic} in agentic, model-forward pipelines and systematically probe VLM slide comprehension.
Our contributions are threefold.
First, we curate a diverse dataset of PowerPoint decks and extract ground-truth geometry, style, and text via a pipeline combining PowerPoint XML with rasterized renders.
Second, we design protocols for low-level fidelity and structured comprehension, including element-wise Hungarian alignment and refinement-relevant probes of judge reliability (variance, sensitivity) and robustness via controlled perturbations to geometry, style, and text.
Third, we extend evaluation to deck-level narrative by asking VLMs to reorder shuffled slides, assessing coherence.

Applying VLM-SlideEval, we surface clear limits and strengths. VLMs struggle with pixel-accurate extraction and show behavioral divergence under controlled perturbations, yet they competently extract structured content on single slides while remaining unreliable for deck-level narrative.
These findings caution against over-reliance on current VLMs for fine-grained slide evaluation and {motivate more calibrated critic-in-the-loop refinement and selection gates for agentic pipelines}.

\begin{figure*} 
  \centering
  \includegraphics[width=\linewidth]{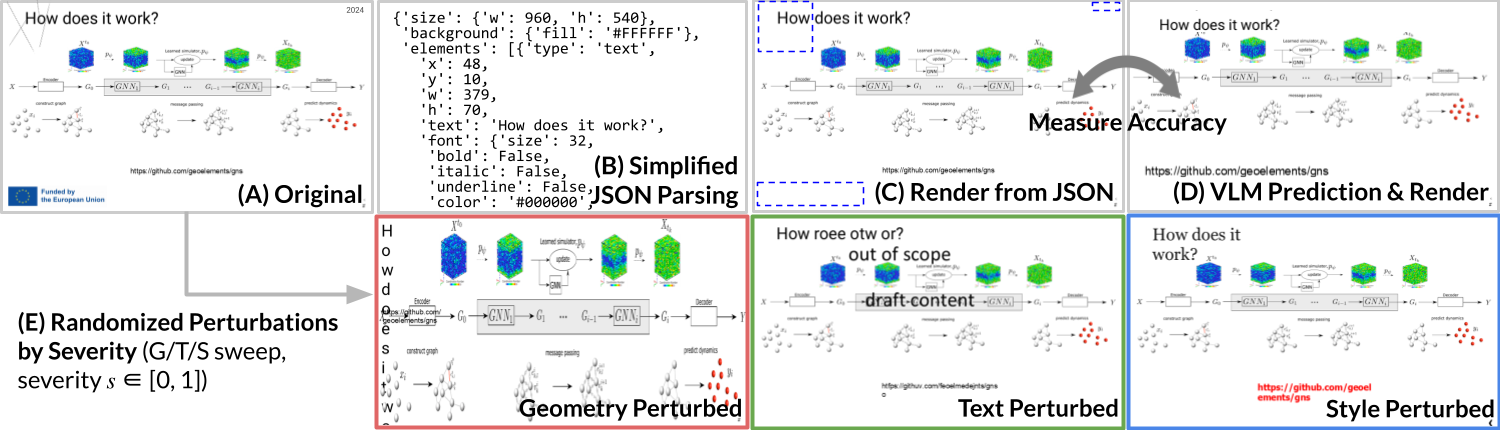}
  \vspace{-1.5em}
  \caption{\small\textbf{Evaluation Task Examples:} Top: From an original slide (A), we parse a simplified schema JSON (Table~\ref{tab:gt_schema}) (B), reconstruct a normalized slide (C; blue dashed boxes show theme-embedded content omitted by the schema). A VLM predicts the schema from the re-rendered slide (D), and we score accuracy.
  Bottom: We subsample 100 decks, retain slides with $\geq 3$ visible elements (234 slides total), and apply perturbations to geometry, text, and style with severity $s\in[0,1]$ (larger $s$ means stronger changes; details in \S\ref{sec:method}). Perturbed slides are then used for VLM quality evaluation and sensitivity analyses (\S\ref{sec:result}).}
  \vspace{-1.5em}
  \label{fig:workflow}
\end{figure*}

%% file: src/02_related_work.tex
\vspace{-1em}
\section{Related Work} \vspace{-.7em}
Calibrated VLM evaluators are increasingly {critical} in agentic, model-forward pipelines: they guide candidate selection, drive iterative refinement at inference time, and even supply reward signals for training.
Recent work shows verifier-guided decoding that improves performance without weight updates~\cite{iad}, generalist multimodal judges used both as LMM-as-a-Judge and as reward models~\cite{llava-critic}, actor-critic loops that critique and correct reasoning~\cite{mmc}, and refinement-centric benchmarks plus standardization frameworks that emphasize granular measurement~\cite{mmrefine,eureka}.
Concurrently, Image2Struct benchmarks VLM image reconstruction on  webpages, LaTex, and musical scores~\cite{roberts2024image2struct}.
This motivates a slide-native, \emph{verifiable} evaluator that produces actionable signals at pixel, element, and deck levels.

Yet VLM evaluation remains challenging. Open-ended judging often relies on incomplete visual context and fuzzy rubrics, yielding inconsistent scores~\cite{prabhu2024trust}, while models hallucinate and make perceptual errors in visually grounded reasoning~\cite{ma2024mmlongbench}.
Under \emph{controlled manipulations} and counterfactuals, VLMs may inject priors unsupported by pixels and show limited sensitivity to fine-grained changes~\cite{guan2024hallusionbench,vo2025visionlanguagemodelsbiased}.
Robustness studies further find text corruptions especially damaging, lightweight adapters sometimes rivaling full fine-tuning, and broader axes (fairness, toxicity, multilinguality) underexplored~\cite{chen2023benchmarking,lee2024vhelm}.

Slide presentations sit within multimodal document understanding, where \emph{structured parsing} underpins both comprehension and authoring.
Prior work has explored language-driven manipulation of slide \emph{objects} (not pixels) for faster, faithful editing~\cite{jung2025talk}, OCR-free pretraining for screenshots and UI/text layouts that improves element-level parsing~\cite{lee2023pix2struct}, and automatic extraction of deck structure for role identification and accessibility~\cite{peng2023slide}.
In parallel, systems that generate slides from long-form documents highlight the need for \emph{scalable, slide-specific evaluation}~\cite{fu2022doc2ppt,ge2025autopresent}.

Unlike work that omits a slide-native evaluator, relies on QA proxies, or focuses robustness on charts/UIs, \emph{VLM-SlideEval} provides a slide-specific framework that couples pixel-accurate alignment to PPT-native ground truth with slide-relevant perturbations and deck ordering, \emph{positioning the evaluator as a critic for agentic pipelines}.

%% file: src/03_method.tex
\vspace{-.5em}
\section{Method} \label{sec:method}
\vspace{-1em}
\noindent\textbf{Data Source.} We sample 100 English-dominant (\(\ge\!70\%\) by \texttt{langid}~\cite{langid}) \texttt{.pptx} decks from Zenodo10K (legacy \texttt{.ppt} excluded), totaling 1{,}948 slides, with CC-BY 4.0 license (Summary statistics in Appendix~\ref{app:gt_extraction}, Table~\ref{tab:gt-summary}).

\noindent\textbf{Ground Truth}
Element \emph{geometry}, \emph{content}, and \emph{style} are extracted from PowerPoint XML and post-layout rendering. We parse static XML and then query the COM (Component Object Model) API after a layout pass to recover effective font metrics and tight text bounds (mitigating AutoFit and container/tight-box discrepancies). Elements are stored in a standardized schema with explicit units (Appendix~\ref{app:gt_extraction}, Table~\ref{tab:gt_schema}).

\noindent\textbf{VLM Parsing \& GT Matching.}
Slides are rasterized to PNG and sent with a fixed 960\(\times\)540px coordinate frame; we test five VLMs (via Azure) to return JSON validated against our schema (invalid JSON counts as a parse failure). Each slide is run ($N=3$) times (low temperature), and metrics are reported per-run and pooled.
Predictions are aligned to GT via Hungarian matching (\cf~\cite{kuhn1955hungarian,Stewart_2016_CVPR,carion2020end,dong2025scan,wang2025infinity}) with a blended cost (1-IoU, center/size difference; text adds content distance) and an acceptance gate; details in Appendix~\ref{app:matching}.

\noindent\textbf{Perturbation Synthesis.}
\emph{Seeds.}
From the same 100 decks we manually select slides well-preserved by the schema and with at least a minimal complexity, \(\ge\!3\) visible text elements, yielding 234 seeds; the reconstructed slide serves as the clean baseline.
\emph{Severity knobs.}
We generate controlled degradations along \emph{geometry}, \emph{text}, and \emph{style}, parameterized by a single severity \(s\in\{0,0.1,\ldots,1.0\}\).
Magnitudes (\eg~pixel offsets, font-size factors) and event probabilities (\eg~drop/insert text boxes) increase monotonically with \(s\); randomness is seeded per (slide, axis, \(s\)).
Exports use a Node.js-based PPTX builder and headless rendering.
From the 7{,}722 original+perturbed slides in total (hyperparameters in App.~\ref{app:perturbations}), we subsample up to 50 slides per severity per axis for evaluation.

\noindent\textbf{Manipulation Check.} We assess whether increasing severity \(s\in[0,1]\) yields orderly and proportional degradation using (i) \emph{adjacent POA} (POA\(_\text{adj}\) := the fraction of consecutive severity steps where \(y^\ast\) does not decrease - and (ii) the \emph{mean absolute calibration error} (MACE) to the identity \(y^\ast = s\), on the normalized \([0,1]\) scale. Empirically, POA is high (5-pt \(\approx 0.95\); 100-pt \(\approx 0.80\)) with moderate calibration (overall MACE \(\approx 0.34\)).


\noindent\textbf{Analysis \& Measures}
We evaluate: (i) \emph{parseability} (schema-valid JSON rate); (ii) end-to-end (e2e) and parsed-only \emph{extraction quality} on matched elements (geometry, content, style); (iii) \emph{narrative ordering} (deck reordering; Kendall's \(\tau\), Spearman's \(\rho\)); and (iv) \emph{perturbation sensitivity} - \(R^2\), POA and \(\operatorname{Spearman}(\text{severity},\, y^\ast)\) - comparing different evaluator scales and models. We report bootstrap 95\% CIs where appropriate. Full metric definitions and evaluator prompts appear in Appendix~\ref{app:analysis_details}.

%% file: src/04_results.tex
\begin{figure*} 
  \vspace{-\baselineskip} 
  \centering
  \includegraphics[width=\linewidth]{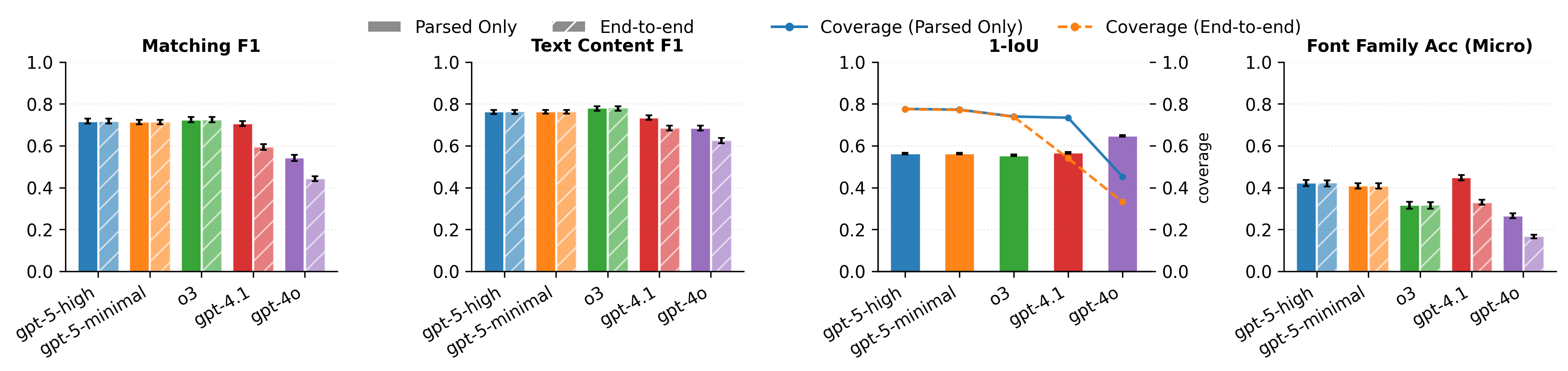}
  \vspace{-2em}
    \caption{
    \small{Parsed-only (solid) vs.\ e2e (hatched) with coverage (\ie~fraction of ground truth instances evaluated for the metric; lines). 
    \texttt{o3}/\texttt{gpt-5} lead on Matching F1 (0.71-0.72) and Text Content F1 (0.76-0.78); 
    \texttt{o3} best in geometry (1-IoU $0.55$).
    Font Family Accuracy is low overall (max $0.42$).
    More results in Fig.~\ref{fig:other_bars}, Appendix~\ref{app:detailed_results}.
    }}
    \vspace{-1.5em}
    \label{fig:headline_bars}
    \end{figure*}

\vspace{-0.5em}
\section{Results}  \label{sec:result}
\vspace{-1em}
We benchmark five VLMs (Azure API) on three main tasks: 1) element-level extraction from slides, 2) behavior under controlled perturbations, and 3) narrative understanding via slide re-ordering. 

\xhdr{Slide Parseability}
Parse success declines with slide complexity for {GPT-4.1} (about $93\%$ for simple slides with $\leq 8$ elements,  $72.1\%$ for (8-16], $32.8\%$ for (16-32], and $18.2\%$ for $\geq 32$ elements). 
{GPT-4o} follows a similar trend but with an earlier decline: about $88.0\%$ for $\leq 8$, $57.6\%$ for (8-16], $45.8\%$ for (16-32], with a small (noisy) uptick to $66.7\%$ at $\geq 32$ ($N = 66$).
In contrast, {o3} and the {GPT-5} variants remain effectively at ceiling across all bins ($99.5\%+$). See Fig.~\ref{fig:parseability_curve}.

\xhdr{Element Prediction Accuracy}
Across headline metrics (Fig.~\ref{fig:headline_bars}), {o3} and the {GPT-5} variants lead under e2e. 
\emph{Matching F1}: Parsed $\rightarrow$ e2e performance drops ($\Delta\approx0.12$ for GPT-4.1 and GPT-4o), with {o3} achieving the highest e2e F1 score ($0.72$), followed by {GPT-5} ($0.71$-$0.72$), vs.\ {GPT-4.1} ($0.59$) and {GPT-4o} ($0.44$). 
\emph{Text Content F1}: {o3} $0.78$ (best), GPT-5 $0.76$, GPT-4.1/GPT-4o $0.69/0.63$. 
\emph{Geometry} (1-IoU; lower better): {o3} (best, $0.55$), GPT-5 ($0.56$), GPT-4.1 ($0.57$), GPT-4o (worst, $0.65$).
E2e coverage is limited, especially for GPT-4o (0.33) and GPT-4.1 (0.54) vs the rest (0.74-0.78) 
\emph{Styling} (Font Family Acc.): overall low ($0.17$-$0.42$), with GPT-5-high highest ($0.42$) and GPT-4o lowest ($0.17$).
Detailed metrics and parsed-only comparisons appear in Table~\ref{tab:extraction-summary} and Fig.~\ref{fig:other_bars} (App.~\S~\ref{appendix_subsection_extraction_performance}).

\xhdr{Behavior Under Controlled Perturbations - Scale correspondence}
\begin{figure}[t]
  \centering
  \begin{subfigure}{0.49\linewidth}
    \centering
    \includegraphics[width=\linewidth,height=\panelheight,keepaspectratio]{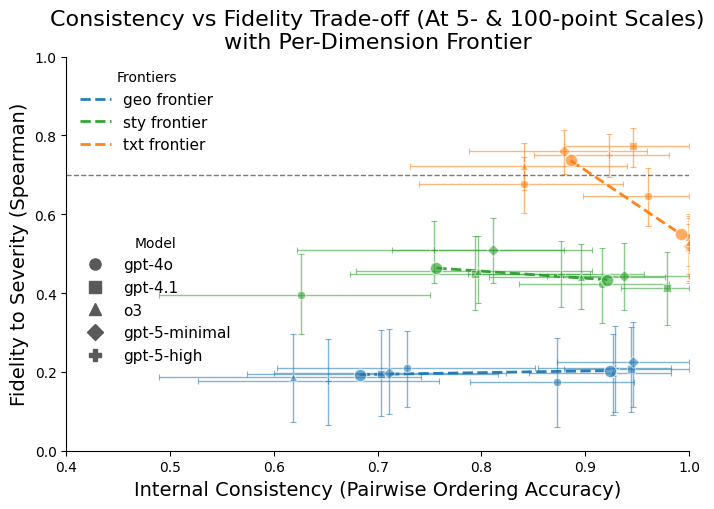}
      \caption{\small\textbf{Consistency-fidelity frontier} per dimension. Consistency is POA\(_\text{adj}\) and fidelity is Spearman \((\text{severity},\,y^\ast)\) (higher better). Geometry/style show no fidelity gain but lower consistency when moving from 5- to 100-pt scale; text trades fidelity with consistency.}
    \label{fig:frontier}
  \end{subfigure}\hfill
  \begin{subfigure}{0.49\linewidth}
    \centering
    \includegraphics[width=\linewidth,height=\panelheight,keepaspectratio]{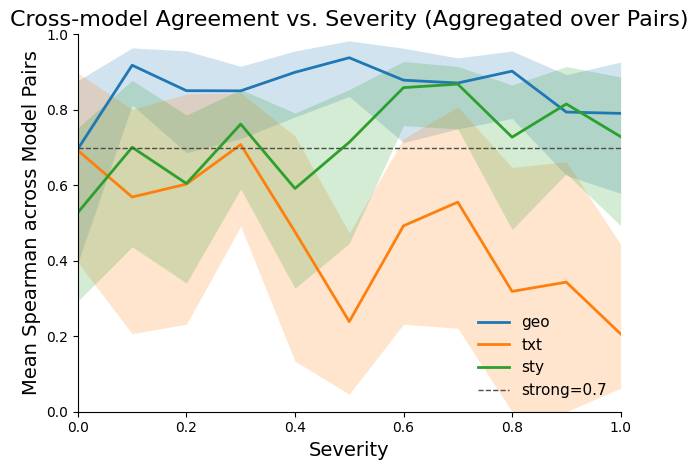}
    \caption{\small\textbf{Cross-model agreement vs.\ severity}. Spearman agreement across model pairs by severity buckets. Geometry/style pairs often exceed 0.80-0.90; text is lowest (best text pair \(\overline{\rho}\approx0.55\)), indicating limited interchangeability on text.}
    \label{fig:agreement}
  \end{subfigure}
  \vspace{-.5em}
  \caption{\small\textbf{Evaluation results of model behavior under controlled perturbations.}}
  \vspace{-1.5em}
  \label{fig:results-panels}
\end{figure}
Within each model, an \emph{isotonic link} maps 5-point scores to 100-point scores with high fidelity: \(R^2 \in [0.85, 0.89]\) across models (\(p = 0.001\)), with GPT-4.1 the tightest (\(\mathrm{RMSE} = 0.075\)) and others close (\eg~GPT-5-high \(0.083\)) on the normalized degradation scale \(y^\ast \in [0,1]\).
This establishes that the two {scales} are largely monotone reparameterizations.
\emph{However}, a monotone mapping does not imply identical behavior under controlled severity shifts: coarse 5-point scores may reduce quantization jitter and improve within-slide ordering, whereas 100-point scores may expose finer variation that can either reflect genuine sensitivity or add noise.
We therefore examine explicit \emph{scale\(\times\)dimension} trade-offs below.

\xhdr{Scale$\times$dimension trade-offs}
We quantify \emph{internal consistency} as POA\(_\text{adj}\) and \emph{fidelity} as \(\operatorname{Spearman}(\text{severity},\, y^\ast)\).
We find that for \textbf{geometry} and \textbf{style}, moving from 5-pt to 100-pt yields \emph{no material fidelity gain} (bootstrap CIs overlap across models) but \emph{reduces} POA\(_\text{adj}\), as implied by the flat frontiers (\eg~$[0.87,0.95]$ $\to$ $[0.62,0.73]$ (geometry); $[0.88,0.98]$ $\to$ $[0.63,0.81]$ (style)) (Fig.~\ref{fig:frontier}).
Thus a coarser scale is preferable for stability in these dimensions.
In contrast, for \textbf{text}, 100-pt \emph{increases fidelity substantially} (\eg~GPT-5-high $0.51 \to 0.75$; GPT-5-minimal $0.52 \to 0.76$) while \emph{lowering} POA\(_\text{adj}\) ($1.00$ $\to$ $[0.88,0.92]$), revealing a consistency-fidelity trade-off.

\xhdr{Model interchangeability}
Models \emph{diverge most on text} (Fig.~\ref{fig:agreement}). Even the most convergent text pair (GPT-5-high vs.\ GPT-5-minimal) attains only $\overline{\rho} \approx 0.55$ (mean of per-severity Spearman), whereas geometry/style pairs frequently exceed $[0.80,0.90]$. 
Notably, the most divergent geometry pair (\eg~GPT-4o vs. o3) still shows higher agreement ($\overline{\rho} \approx 0.78$) than the most convergent text pair, underscoring that \emph{text quality} is the most divergent axis for cross-model agreement.

\xhdr{Narrative in Slide Deck}
Overall (Figure \ref{fig:ordering}), the models exhibit difficulty in accurately predicting slide order, with Kendall's $\tau \in [0.04,0.12]$, Spearman's $\rho \in [0.05,0.13]$, and Exact Match scores $s \in [0.10,0.17]$) only marginally outperforming random guessing, yet remaining below the theoretical upper bound of 1.0.
This suggests that the models may struggle to comprehend and reason through the narrative flow of a presentation. 
Among them, GPT-4.1 delivered the strongest performance ($[0.04,0.07]$ point of improvement) over GPT-5-minimal (Details in Appendix~\ref{app:ordering}).

%% file: src/05_conclusion.tex
\vspace{-1em}
\section{Conclusion} \vspace{-1em}
We present \textbf{VLM-SlideEval}, a framework for evaluating slide element extraction, robustness to controlled perturbations, and narrative reordering on a curated PPTX corpus with ground truth.
Newer VLMs (o3, GPT-5) outperform GPT-4.1/GPT-4o, yet all struggle with pixel-accurate style (\eg~fonts) and cross-slide narrative coherence, and under perturbations exhibit a fidelity-consistency trade-off: geometry/style are comparatively stable, while finer text scales raise sensitivity but reduce internal score consistency.
These findings argue for calibrated, slide-native evaluator in agentic/model-forward pipelines, using verifiable and accurate signals to gate selection and steer iterative refinement.
Limitations include public PPTX, seeded perturbations, the suite of VLMs evaluated, as well as the simplified schema used for parsing slides; future work spans broader corpora, richer narrative probes, stronger verifiable checks, and judge calibration.

%% file: src/06_appendix.tex
\appendix

\section{Ground Truth Extraction Details}
\label{app:gt_extraction}
Ground truth elements are obtained by parsing the PowerPoint XML specification and cross-checking against a PNG export of the same slides.
Each element type (text, rect, line, image, table) is represented in a unified schema with pixel-based geometry and absolute units for fonts and strokes (the full extraction schema is shown in Table~\ref{tab:gt_schema} below).

\begin{table}[h]
\centering
\small
\begin{tabular}{lll}
\toprule
\textbf{Field(s)} & \textbf{Applies to} & \textbf{Unit / Notes} \\
\midrule
\texttt{w}, \texttt{h} & slide & px; {fixed at 960$\times$540} \\
\texttt{x}, \texttt{y}, \texttt{w}, \texttt{h} & rect, text, image, table & px; top-left anchor \\
\texttt{x1}, \texttt{y1}, \texttt{x2}, \texttt{y2} & line & px; line endpoints \\
\texttt{rx} & rect & px; corner radius \\
\texttt{strokeWidth} & rect, line & points (pt); absolute width \\
\texttt{font.size} & text & pt; absolute font size \\
\texttt{font.style} & text & categorical; bold, italic, underscore \\
\texttt{color fields} & text, slide, line, rect & normalized hex (\#RRGGBB) \\
\texttt{align} & text & categorical; left/center/right/justify/distributed \\
\bottomrule
\end{tabular}
\caption{Schema of extracted ground truth fields (excerpt). See Appendix~\ref{app:gt_extraction} for full details.}
\label{tab:gt_schema}
\end{table}

We normalized the coordinates to the fixed slide size 960×540px, with its origin at the top-left corner.
For styling information, font sizes are reported in points, while color values are normalized into \#RRGGBB format.
This enables precise cross-comparison between extracted ground truth and predictions returned by vision-language models (see Sec.~\ref{sec:method}).
The summary statistics of ground truth element extraction can be found in Table~\ref{tab:gt-summary}
\begin{table}[t]
\small
\setlength{\tabcolsep}{3pt}
\centering
\begin{tabular*}{\linewidth}{@{\extracolsep{\fill}} lrrrrr rrrrr r @{}}
\toprule
& \multicolumn{5}{c}{\textbf{Per deck}} & \multicolumn{5}{c}{\textbf{Per slide}} & \textbf{Total} \\
\cmidrule(lr){2-6}\cmidrule(lr){7-11}
\textbf{Category} & \textbf{Mean} & \textbf{SD} & \textbf{Min} & \textbf{Med} & \textbf{Max} & \textbf{Mean} & \textbf{SD} & \textbf{Min} & \textbf{Med} & \textbf{Max} & \textbf{Sum} \\
\midrule
Num.\ of slides                & 19.48 & 11.54 & 1   & 18.0 & 46   & ---  & ---  & --- & --- & --- & 1948 \\
\addlinespace[2pt]
All elements                   & 119.01 & 142.07 & 1  & 93.0 & 1183 & 6.11 & 9.03 & 0 & 4.0 & 153 & 11901 \\
\addlinespace[2pt]
\textit{By type} \\
\quad Text                     & 63.40  & 58.40  & 0  & 49.0 & 314  & 3.25 & 3.34 & 0 & 3.0 & 69 & 6340 \\
\quad Rect                     & 15.44  & 63.66  & 0  &  2.5 & 622  & 0.79 & 5.28 & 0 & 0.0 & 93 & 1544 \\
\quad Line                     &  5.64  & 18.74  & 0  &  0.0 & 167  & 0.29 & 2.12 & 0 & 0.0 & 49 & 564 \\
\quad Image                    & 33.71  & 33.50  & 0  & 28.0 & 172  & 1.73 & 2.54 & 0 & 1.0 & 44 & 3371 \\
\quad Table                    &  0.82  &  4.09  & 0  &  0.0 &  40  & 0.04 & 0.35 & 0 & 0.0 & 11 & 82 \\
\bottomrule
\end{tabular*}
\caption{Ground-truth extraction summary across 100 decks and 1{,}948 slides. Per-deck statistics are computed across decks; per-slide statistics across slides.
}
\label{tab:gt-summary}
\end{table}

\section{Predicted Extraction Prompt}
\label{app:prediction_prompt}
\begin{figure*}[h]
\centering

\begin{lstlisting}[style=prompt]
(*@\noindent\color{black}\rule{\linewidth}{0.5pt}@*)
[System Message]
Analyze the location, size, and styling information of elements in the slide.
The size of the slide is: {TARGET_W} (w) x {TARGET_H} (h) pixels. The screenshot of the slide was taken at DPI = 72.
Top-left of the slide is (0,0), +x rightward, +y downward.
All geometry fields are integers in pixels, unless noted otherwise.

Return a JSON object with the following top-level fields for the single slide:
{ size, background, texts:[], rects:[], lines:[], images:[], tables:[] }.
Include every required field exactly as specified.
\end{lstlisting}

\vspace{0.75em}

{%
\{\emph{Extraction Specification Information: Table~\ref{tab:gt_schema} Content Here}\}
}%

\vspace{0.75em}

\begin{lstlisting}[style=prompt]
[User Message]
{"type": "image_url", "image_url": {"url": "<base64_thumbnail>", "detail": "auto"}}
(*@\noindent\color{black}\rule{\linewidth}{0.5pt}@*)
\end{lstlisting} \vspace{-1em}
\caption{Prompt used for structured extractions from VLMs for a single slide image.}
\label{fig:extraction_prompt}
\end{figure*}
We use a single-slide prompt that (i) fixes the slide coordinate frame at 960×540px with origin at the top-left; (ii) specifies units per field (pixels for geometry, points for fonts and strokes, hex for colors); and (iii) enumerates the required output schema ({size, background, texts, rects, lines, images, tables}) with field-level guidance (\eg~x,y are the top-left of the element bbox; lines use x1,y1,x2,y2; rectangle corner radius is rx).
The system message instructs the VLM to return a strict JSON object for the single image provided.
A compact reference table in the prompt reiterates allowed values (\eg~text align $\in$ \{left, center, right, justify, distributed\}) and clarifies that font and stroke widths are in points (absolute), while all positions and sizes are in pixels.
The slide image is passed inline as a base64 PNG.
We enforce structured output via the API's JSON schema mode and validate responses with Pydantic; invalid JSON or schema mismatches are marked as parse failures.

\begin{algorithm}[H]
\caption{Hungarian Matching with Blended Geometry+Content Cost and Threshold Gate}
\label{algo:hungarian_pseudo_code}
\begin{algorithmic}[1]
\State \textbf{Input:} $G=\{g_i\}_{i=1}^{m}$, $P=\{p_j\}_{j=1}^{n}$
\State \textbf{Params:} slide size $(W,H)$; weights $(\alpha,\beta,\gamma,\delta)$; blended acceptance threshold $\tau \in [0,1]$
\State \textbf{Accessors:} $\text{box}(e)\!\rightarrow\!(x,y,w,h)$; $\text{sim}(g,p)\in[0,1]$ if available (else set $\delta{=}0$)
\vspace{0.25em}
\State \textbf{Defs:}
\State $\mathrm{IoU}(a,b)=\frac{\mathrm{area}(a\cap b)}{\mathrm{area}(a)+\mathrm{area}(b)-\mathrm{area}(a\cap b)}$
\State $d_{\mathrm{center}}(a,b)=\frac{\lVert c(a)-c(b)\rVert_2}{\sqrt{W^2+H^2}}$ where $c(\cdot)$ is box center
\State $\mathrm{size\_rel}(a,b)=\tfrac{1}{2}\!\left(\frac{|w_a-w_b|}{\max(\varepsilon,w_a)}+\frac{|h_a-h_b|}{\max(\varepsilon,h_a)}\right)$
\vspace{0.25em}
\State Construct $C\in\mathbb{R}^{m\times n}$
\For{$i=1$ to $m$}
  \For{$j=1$ to $n$}
    \State $a\!\gets\!\text{box}(g_i)$,\quad $b\!\gets\!\text{box}(p_j)$
    \State $c_{\text{iou}}\!\gets\!1-\mathrm{IoU}(a,b)$;\quad $c_{\text{center}}\!\gets\!d_{\mathrm{center}}(a,b)$;\quad $c_{\text{size}}\!\gets\!\mathrm{size\_rel}(a,b)$
    \State $c_{\text{cont}}\!\gets\!1-\text{sim}(g_i,p_j)$ \textbf{if} content available \textbf{else} $0$
    \State $C_{ij}\!\gets\!\alpha c_{\text{iou}}+\beta c_{\text{center}}+\gamma c_{\text{size}}+\delta c_{\text{cont}}$
  \EndFor
\EndFor
\State Compute optimal assignment $\mathcal{A}\subseteq\{1..m\}\!\times\!\{1..n\}$ by Hungarian on $C$
\vspace{0.25em}
\State \textbf{Threshold gate and bookkeeping}
\State $\mathcal{M}\!\gets\!\varnothing$;\ \ $\text{matchedG}\!\gets\!\varnothing$;\ \ $\text{matchedP}\!\gets\!\varnothing$
\For{each $(i,j)\in\mathcal{A}$}
  \If{$C_{ij}\le\tau$}
     \State $\mathcal{M}\!\gets\!\mathcal{M}\cup\{(i,j)\}$;\ \ $\text{matchedG}\!\gets\!\text{matchedG}\cup\{i\}$;\ \ $\text{matchedP}\!\gets\!\text{matchedP}\cup\{j\}$
  \EndIf
\EndFor
\State \textbf{Output:} matches `$\mathcal{M}$', false positives `$P\setminus\text{matchedP}$', false negatives `$G\setminus\text{matchedG}$'
\end{algorithmic}
\end{algorithm}

\section{Prediction-to-Ground Truth Matching Algorithm} \label{app:matching}
Let $G=\{g_i\}$ denote the set of ground truth elements and $P=\{p_j\}$ the predicted elements.
Each candidate match $(g_i, p_j)$ ($c_{ij} \in C \in \mathbb{R}^{|G|\times|P|}$) we define a blended cost $c_{ij} = \alpha \bigl(1 - \mathrm{IoU}(g_i, p_j)\bigr)
       + \beta \, d_{\text{center}}(g_i, p_j)
       + \gamma \, \mathrm{size\_rel}(g_i, p_j)
       + \delta \, \bigl(1 - \mathrm{sim}(g_i, p_j)\bigr),$
where $\mathrm{IoU}$ is the box overlap, $d_{\text{center}}$ is normalized Euclidean center distance, $\mathrm{size\_rel}$ is relative size drift, and $\mathrm{sim}$ is a content similarity score (\eg~normalized text similarity). 
We solve a minimum-cost bipartite matching with the Hungarian algorithm~\cite{kuhn1955hungarian,carion2020end} on $C=[c_{ij}]$.
Finally, we apply a lightweight sanity check: a matched pair $(i,j)$ is accepted iff its blended cost is below a threshold $\tau$ (\ie~$c_{ij}\le\tau$); otherwise it is discarded, yielding an unmatched ground-truth (FN) and prediction (FP).
Pseudo code of this procedure can be found in Algorithm~\ref{algo:hungarian_pseudo_code}.

This formulation generalizes naturally to other modalities; only the similarity term $\text{sim}(\cdot)$ is type-dependent.
For example, table elements may use cell-value overlap, and images may use caption, color histogram, and object-scene similarity.

\section{Perturbation Operators and Hyperparameters}
\label{app:perturbations}

\xhdr{Notation} We perturb a slide's element list \(\mathcal{E}\) with a single strength knob \(s\in[0,1]\).
When \(s=0\) the transform is a no-op (we return a deep copy).
All probabilities and noise scales below are monotone in \(s\), and all randomness is seeded for reproducibility.

\xhdr{Geometry (layout/alignment)} We act on ``box-like'' elements with geometry \((x,y,w,h)\) (text, image, table, rect, chart).
For each eligible element (sampled with per-element probability \(\pi_{\mathrm{geo}}\); default \(=1.0\)):

\begin{itemize}\itemsep1pt
\item\textbf{Translation:} \((x',y')=(x+\Delta_x,\;y+\Delta_y)\) with \(\Delta_x\sim\mathcal{N}(0,\sigma_x^2)\), \(\Delta_y\sim\mathcal{N}(0,\sigma_y^2)\),
\[
\sigma_x(s) = (0.04+0.16\,s)\cdot W,\quad \sigma_y(s) = (0.04+0.16\,s)\cdot H,
\]
where \((W,H)\) is slide size (960\(\times\)540px).
\item \textbf{Scaling:} \((w',h')=(w\cdot \eta_w,\;h\cdot \eta_h)\), with \(\eta_{\{\cdot\}}\sim \exp(\mathcal{N}(0,\sigma_{\log}^2))\) and \(\sigma_{\log}(s)=0.12+0.55\,s\).
\item \textbf{Extreme size (optional):} with probability \(p_{\mathrm{ext}}(s)=0.20\,s\), additionally multiply \((w',h')\) by
\[
r \sim \mathrm{Uniform}(0.15,0.50)\;\;\text{or}\;\;\mathrm{Uniform}(1.5,10).
\]
\item \textbf{Reposition (optional):} with probability \(p_{\mathrm{rep}}(s)=0.10\,s\),
sample a fresh \((x',y')\) uniformly over valid canvas positions (respecting current size).
\item \textbf{Collapse (optional):} with probability \(p_{\mathrm{col}}(s)=0.08\,s\),
set one dimension to \(\mathrm{Uniform}(1,3)\) px (skinny or flat).
\item \textbf{Bounds:} clamp to \([0,W-w']\times[0,H-h']\) unless \texttt{allow\_clipping}.
\end{itemize}

\xhdr{Text Content}
We operate on text elements; non-text are passed through.
For each text box (sampled with per-element probability \(\pi_{\mathrm{txt}}\); default \(=1.0\)):

\begin{itemize}\itemsep2pt
\item \textbf{Character-level noise} with per-character rate
\(p_{\mathrm{char}}(s)=p_{\min}+(p_{\max}-p_{\min})\,s\),
where \(p_{\min}=0.02\), \(p_{\max}=0.25\). For each affected character, apply one of
\(\{\)substitute, delete, insert, adjacent-swap\(\}\) with weights \((0.50,0.20,0.15,0.15)\).
Substitutions/insertions prefer keyboard-neighbor letters; case preserved.
\item \textbf{Numeric preservation (optional):} after noise, restore the original numeric runs
(\verb|\d+(\.\d+)?|) in textual order to limit semantic drift on quantities.
\item \textbf{Drop boxes (optional):} with probability \(p_{\mathrm{drop}}(s)=0.18\,s\), remove the entire text box.
\item \textbf{Insert boxes (optional):} with probability \(p_{\mathrm{ins}}(s)=0.35\,s\),
insert \(n\in\{1,\ldots,\min(\texttt{max\_inserts},\,1+\lfloor 3s\rfloor)\}\) irrelevant text boxes.
Each insertion samples geometry fractions
\(w/W\sim\mathrm{U}(0.15,\,0.35+0.35s)\), \(h/H\sim\mathrm{U}(0.08,\,0.22+0.28s)\),
with uniform valid \((x,y)\). Text is drawn from a small pool (\eg~``lorem ipsum'', ``TODO: revise''), and default font attributes are assigned (size scales with \(s\); emphasis toggles with small \(s\)-scaled probabilities).
\end{itemize}

\xhdr{Style (typography \& color)}
We act on text elements (per-element probability \(\pi_{\mathrm{sty}}\); default \(=1.0\)).
Let \(f\) denote a font object with fields \{\texttt{name}, \texttt{size}, \texttt{bold}, \texttt{italic}, \texttt{underline}, \texttt{color}\}.

\begin{itemize}\itemsep2pt
\item \textbf{Family switch:} with probability \(p_{\mathrm{fam}}(s)=0.20+0.60\,s\),
replace \texttt{name} by a random choice from a fixed pool excluding the current family.
\item \textbf{Size jitter:} \(\texttt{size}'=\mathrm{clip}_{[6,120]}\big(\texttt{size}\cdot \exp(\mathcal{N}(0,\sigma_{\mathrm{sz}}^2))\big)\)
with \(\sigma_{\mathrm{sz}}(s)=0.45\,s\).
With probability \(p_{\mathrm{szext}}(s)=0.25\,s\), additionally multiply by
\(\mathrm{U}(0.12,3.8)\) to produce tiny/huge outliers.
\item \textbf{Emphasis toggles:} independently flip \{\texttt{bold}, \texttt{italic}, \texttt{underline}\} with probability \(p_{\mathrm{tog}}(s)=0.20\,s\).
\item \textbf{Color:} with probability \(p_{\mathrm{inj}}(s)=0.30\,s\), inject an incongruent palette color
(e.g., \#FF0000, \#FFFF00, \#00FFFF, …).
Otherwise jitter the current color in HLS:
\(\Delta h\sim \mathrm{U}(-30^\circ,30^\circ)\,s\),
\(\Delta \ell\sim \mathrm{U}(-0.25,0.25)\,s\),
\(\Delta s\sim \mathrm{U}(-0.20,0.20)\,s\).
With probability \(p_{\mathrm{lowc}}(s)=0.25\,s\), move toward the background color by
\(c'=(1-\alpha)c+\alpha\,c_{\mathrm{bg}}\) with \(\alpha=0.25+0.65\,s\).
\item \textbf{Background:} with probability \(p_{\mathrm{bg}}(s)=0.20\,s\), jitter the slide background color as above.
\end{itemize}

\section{Additional Details for Analysis \& Measures}
\label{app:analysis_details}

\subsection{Slide Parseability}
\label{app:parseability}
\xhdr{Definition} A slide is counted as \emph{parsed} if the model returns a JSON object that validates against our strict schema (fields, types, units) using Pydantic. Responses that are not valid JSON or violate the schema are marked as failures. Parseability is independent of matching quality (later we report on both the end-to-end - including parse failure cases where they would count towards the denominators of the downstream performance metrics - as well as the parsed-only - excluding parse failure cases from analysis; see Fig.~\ref{fig:headline_bars} and Fig.~\ref{fig:other_bars} for the relevant results).

\xhdr{Complexity} We use GT scene complexity $c$ as the total number of ground truth elements on a slide (sum over text, image, table, line, rect, table).


\xhdr{Reliability curve by complexity} Let $\{B_k\}$ be $K$ quantile bins of $c$. For each bin $B_k$ we report
\[
\widehat{\Pr}(\text{success}\mid c\!\in\!B_k) \;=\; \frac{1}{|B_k|}\sum_{i\in B_k}\mathbb{1}\{\text{parsed}_i\},
\]
with a 95\% bootstrap confidence interval via percentile or BCa intervals.

\subsection{Metric Definitions}
\label{app:metrics}

To investigate the VLM slide comprehension accuracy, we measure a suite of metrics encompassing a diverse set of elements for the three dimensions of quality, as detailed below.

\paragraph{Matching counts \& PRF1.}
For each family and overall (micro), precision $\mathrm{P} = \frac{TP}{TP+FP}$, recall $\mathrm{R}=\frac{TP}{TP+FN}$, and $\mathrm{F1}=\frac{2\mathrm{PR}}{\mathrm{P}+\mathrm{R}}$.

\paragraph{Geometry terms (interpretable).}
For boxes we report: $1-\mathrm{IoU}$; center distance $d^{\mathrm{center}}$; relative size $r^{\mathrm{size}}$; for images, aspect-ratio error $r^{\mathrm{ar}}$; for rectangles, radius error $r^{\mathrm{rx}}$; for lines, relative length error $r^{\mathrm{len}}$ and angular error $r^{\mathrm{ang}}$. All terms are in $[0,1]$ after normalization. Lower is better.

\paragraph{Content similarity.}
Text strings are normalized by lowercasing, replacing “\&$\to$and”, stripping punctuation, and collapsing whitespace. We compute $s^{\mathrm{content}}=\texttt{SequenceMatcher}(\tilde{t}_\text{pred}, \tilde{t}_\text{gt})\in[0,1]$ and also report $1-s^{\mathrm{content}}$ where an error term is desired. (Embedding-based similarity is possible but not used in our primary results.)

\paragraph{Style.}
We measure color differences using CIEDE2000 ($\Delta E_{00}$) computed in CIE $L^*a^*b^*$ space after sRGB$\rightarrow$Lab conversion (D65; $k_L=k_C=k_H=1$). Lower is better.
\emph{Rule-of-thumb}: $\Delta E_{00}\!\lesssim\!0.5$ imperceptible, $0.5\!-\!1$ barely perceptible, $1\!-\!2$ small but visible, $2\!-\!3.5$ clearly noticeable under typical viewing. We evaluate: (i) slide background vs.\ GT; (ii) per element type—font color (\texttt{text}), fill and stroke (\texttt{rect}), and stroke (\texttt{line}).
For numeric style fields we report absolute errors in native units: font size (pt) and stroke width (pt).
For booleans we report mismatch rates (0/1): bold, italic, underline (for \texttt{text}).
All statistics are summarized overall and per type using means, standard deviations, and counts; micro-averaged PRF1 is computed from summed TP/FP/FN.

\paragraph{Aggregation.}
We micro-average PRF1 by summing TP/FP/FN over all slides and runs. For scalar errors we report $\{\text{mean},\,\text{stdev},\,n\}$ over all matched pairs (overall and by type). Where noted, we compute bootstrap 95\% CIs (2{,}000 resamples). Deck-level summaries aggregate per slide, then pool across decks (pooled mean/stdev with sample-size weights).

\paragraph{Units \& coordinate frame.}
All geometry is in a fixed 960×540px slide frame, with stroke width and font size in points.
The rasterization is for screenshots only and does not alter the target coordinate system.

\subsection{Evaluator Prompts}
The prompts used by VLMs for assessing the quality of perturbed slides along text, geometry, and style dimensions are instantiated using a common prompt template (Fig.~\ref{fig:evaluator_prompt}).
For \emph{dimension}, we use \{``text quality'', ``layout geometry'', ``style''\}; we provide two scale set points \{(1,5), (1,100)\} and corresponding \emph{mid-point} as the mean of the end-points, and labels as \{``very poor'', ``acceptable'', ``excellent''\}.
The \emph{how to judge} constraints in each dimension are shown in Table~\ref{tab:how_to_judge} below:
\begin{table}[h!]
\centering
\setlength{\tabcolsep}{6pt}
\renewcommand{\arraystretch}{1.05}
\begin{tabular}{p{0.31\linewidth}|p{0.31\linewidth}|p{0.31\linewidth}}
\hline
\textbf{Text quality} & \textbf{Layout geometry} & \textbf{Style} \\
\hline
\begin{minipage}[t]{\linewidth}\vspace{0pt}
\begin{itemize}\setlength\itemsep{0.2ex}\setlength\parsep{0pt}\setlength\topsep{0.2ex}
\item Clarity and plain language
\item Grammar/spelling
\item Bullet length (prefer one line)
\item Concision (avoid fluff)
\end{itemize}
\end{minipage}
&
\begin{minipage}[t]{\linewidth}\vspace{0pt}
\begin{itemize}\setlength\itemsep{0.2ex}\setlength\parsep{0pt}\setlength\topsep{0.2ex}
\item Alignment to grid/edges/baselines
\item Consistent spacing and margins
\item Balance and visual hierarchy
\item Element sizing matches importance
\end{itemize}
\end{minipage}
&
\begin{minipage}[t]{\linewidth}\vspace{0pt}
\begin{itemize}\setlength\itemsep{0.2ex}\setlength\parsep{0pt}\setlength\topsep{0.2ex}
\item Font family consistency and readability
\item Font size appropriate for viewing distance
\item Contrast and color harmony
\item Consistent emphasis (bold/italic/underline sparingly)
\end{itemize}
\end{minipage}
\\
\hline
\end{tabular}
\caption{\small{\emph{How to judge} constraints used by evaluators.}}
\label{tab:how_to_judge}
\vspace{-0.8em}
\end{table}

\label{app:evaluator_prompts}

\begin{figure*}[t]
\centering
\begin{lstlisting}[style=prompt]
(*@\noindent\color{black}\rule{\linewidth}{0.5pt}@*)
[System Message]
[Role]
You score the {DIMENSION} of a PowerPoint slide.

[Scale]
Return ONE integer on the scale {SCALE_MIN}..{SCALE_MAX} (inclusive).
Anchors:
- Min ({SCALE_MIN}): "{LOW_LABEL}".
{OPTIONAL_MID}- Mid ({SCALE_MID}): "{MID_LABEL}".
- Max ({SCALE_MAX}): "{HIGH_LABEL}".

[How to judge]
Consider only:
{CRITERIA_BULLETS}
(*@\noindent\color{black}\rule{\linewidth}{0.5pt}@*)
[User Message]
{"type": "image_url", "image_url": {"url": "<base64_thumbnail>", "detail": "auto"}}
(*@\noindent\color{black}\rule{\linewidth}{0.5pt}@*)
\end{lstlisting} \vspace{-1em}
\caption{\small{Prompt template used by VLM evaluators on perturbed slides.}}
\label{fig:evaluator_prompt}
\end{figure*}

\section{Detailed Results} \label{app:detailed_results}
\subsection{Slide Parsing Success Rate Conditioned on Scene Complexity}
\noindent\textit{Parseability vs.\ complexity.} Figure~\ref{fig:parseability_curve} visualizes trends across complexity bins; the per-bin summaries are:
\begin{figure*}[h] 
  \centering
  \includegraphics[width=.5\linewidth]{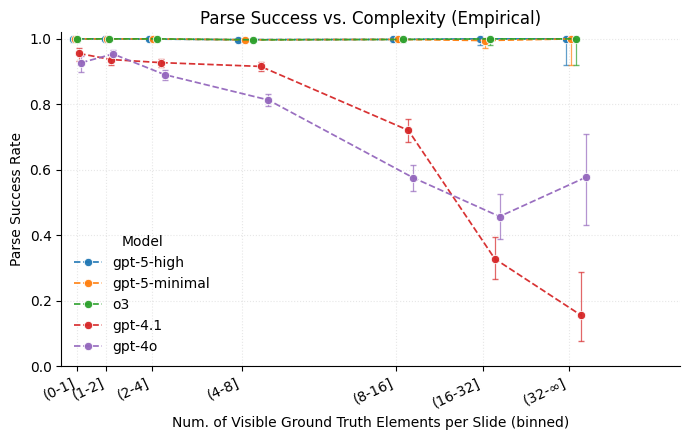}
  \vspace{-.5em}
  \caption{\small{Parse success versus scene complexity (elements per slide) across VLMs.
Complexity bins: $(0,1]$, $(1,2]$, $(2,4]$, $(4,8]$, $(8,16]$, $(16,32]$, $(32,\infty]$.
GPT-5 and {o3} remain near ceiling across bins, while GPT-4 series degrades with complexity.
Estimates in the rightmost bin use small samples ($N = 66$ per model).}}
  \label{fig:parseability_curve}
\end{figure*}
\begin{itemize}
\item \textbf{GPT-5-high} is essentially at ceiling across all complexity bins: five bins are at $\mathbf{100\%}$ and the remaining two are $\mathbf{99.8\%}$–$\mathbf{99.9\%}$.
\item \textbf{GPT-5-minimal} is likewise near-ceiling: $\mathbf{99.7\%}$–$\mathbf{100\%}$ in all but one bin; the lowest bin is $\mathbf{99.5\%}$ (16–32).
\item \textbf{o3} remains at or near ceiling throughout, with $\mathbf{99.7\%}$–$\mathbf{100\%}$ across all bins.
\item \textbf{GPT-4.1} shows clear sensitivity to complexity: $\mathbf{95.5\%}$ (0–1), $\mathbf{93.7\%}$ (1–2), $\mathbf{92.8\%}$ (2–4), $\mathbf{91.6\%}$ (4–8), then drops to $\mathbf{72.1\%}$ (8–16), $\mathbf{32.8\%}$ (16–32), and $\mathbf{18.2\%}$ (32–$\infty$).
\item \textbf{GPT-4o} underperforms \texttt{GPT-4.1} in most bins as complexity grows: $\mathbf{92.7\%}$ $(0,1]$ $\mathbf{95.4\%}$ $(1,2]$, $\mathbf{89.1\%}$ $(2,4]$, $\mathbf{81.4\%}$ $(4,8]$, $\mathbf{57.6\%}$ $(8,16]$, $\mathbf{45.8\%}$ $(16,32]$; the uptick to $\mathbf{66.7\%}$ in $(32,\infty]$ reflects small-sample volatility ($N = 66$).
\end{itemize}

Small sample sizes in the extreme tail $(32,\infty]$, $N = 66$ per model) limit certainty there; the overall pattern is near-perfect parseability for the GPT-5 and {o3} models, with sharp degradation for the GPT-4 series as complexity increases.

\subsection{Extraction Performance} \label{appendix_subsection_extraction_performance}
\begin{figure*}[h] 
    \vspace{-\baselineskip} 
    \centering
    \includegraphics[width=\linewidth]{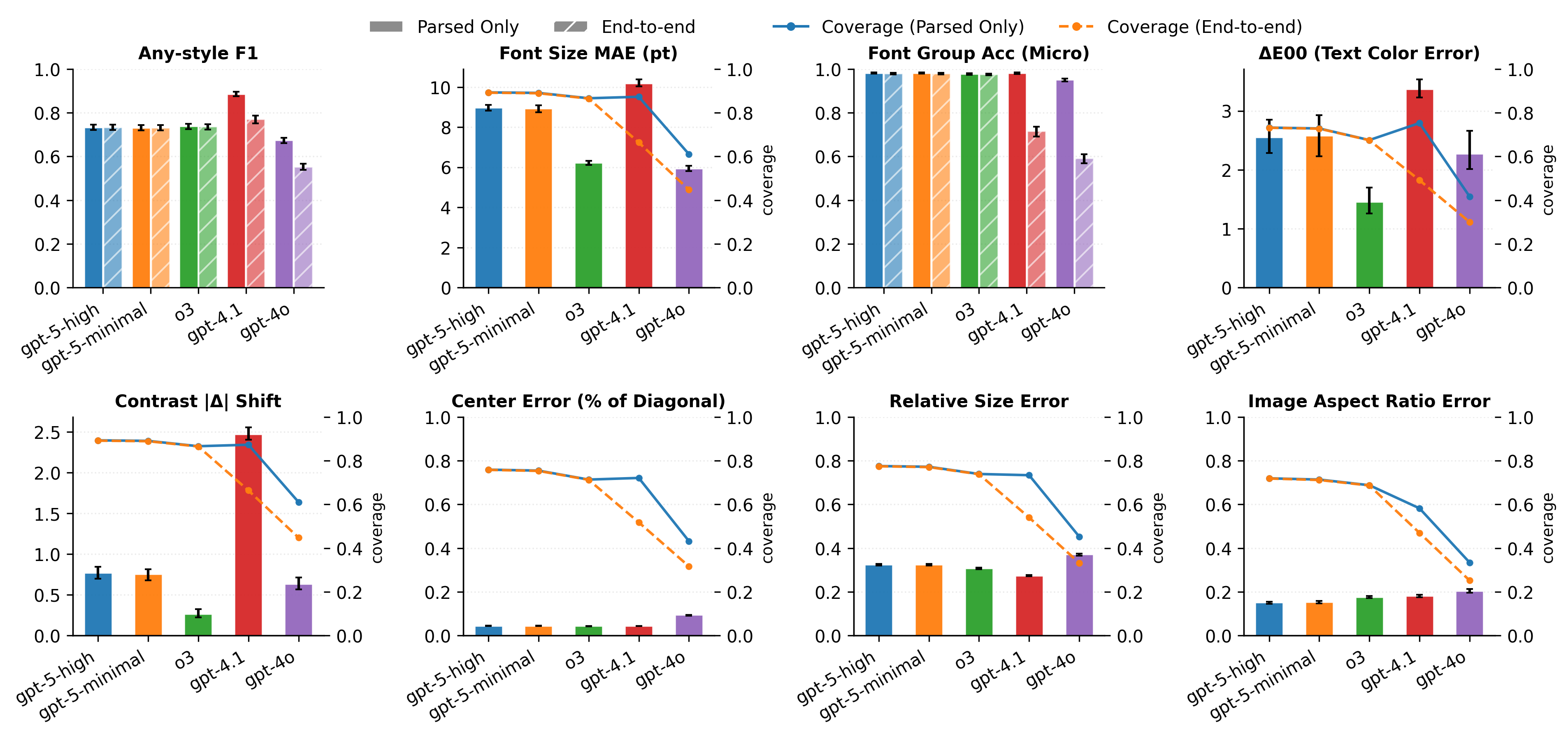}
    \vspace{-1.5em}
    \caption{\small{Bars show \emph{Parsed Only} (solid) vs.\ \emph{End-to-end} (hatched); lines (right axis) show \textbf{coverage} (fraction of ground-truth instances evaluated per metric). 
    \textbf{Styling} (higher is better): Any-style F1 is moderate overall, with {GPT-4.1} at {0.77} (best) and {GPT-4o} at {0.55} (worst); parsed-only boosts are pronounced for the 4-series (\eg~0.89 for {GPT-4.1}, 0.67 for {GPT-4o}). 
    \textbf{Fonts}: font \emph{group} accuracy is near-perfect for {GPT-5-\{minimal,high\}} and {o3} ($\geq\!0.98$) but lower for {GPT-4.1}/{GPT-4o} ($\approx\!0.72/0.59$); font \emph{family} accuracy is substantially lower across models $[0.17,0.42]$. 
    \textbf{Font size}: MAE (pt; lower is better) ranges $[5.93,10.18]$ with {GPT-4o} best. 
    \textbf{Color} (lower is better): text $\Delta E_{00}$ spans $[1.46,3.37]$ ({o3} best, {GPT-4.1} worst) and contrast $|\Delta|$ shift spans $[0.26,2.47]$ ({o3} best, GPT-4.1 worst). 
    \textbf{Geometry} (lower is better): $1-\mathrm{IoU}$ is best for {o3} ({0.55}) and worst for GPT-4o (0.65); center error is $[0.04,0.09]$, size error $[0.27,0.37]$, and image aspect-ratio error $[0.15,0.20]$. 
    End-to-end coverage is substantially lower for the 4-series than for {o3}/GPT-5.}}
    \label{fig:other_bars}
    \vspace{-1.5em}
\end{figure*}

\noindent Fig.~\ref{fig:other_bars} summarizes extraction accuracy and geometry error with \emph{Parsed Only} vs.\ \emph{End-to-end} bars and coverage lines; Table~\ref{tab:extraction-summary} lists per-model metrics, showing e2e (parsed-only) in each cell with best e2e bolded. Overall, {o3} and {GPT-5-\{minimal,high\}} lead across F1/accuracy and geometry, while {GPT-4.1}/{GPT-4o} degrade more under e2e, consistent with lower coverage.

\begin{table*}[h]
\small
\setlength{\tabcolsep}{6pt}
\centering
\resizebox{\textwidth}{!}{%
\begin{tabular}{llllll}
\toprule
\textbf{Metric} & \textbf{GPT-4o} & \textbf{GPT-4.1} & \textbf{o3} & \textbf{GPT-5-minimal} & \textbf{GPT-5-high} \\
\midrule
Element Matching F1 & 0.44 (0.54) & 0.59 (0.71) & \textbf{0.72} (0.72) & 0.71 (0.71) & 0.72 (0.72) \\
\addlinespace[2pt]
\multicolumn{6}{c}{\textbf{Geometry} (micro; lower is better)} \\
\addlinespace[2pt]
$1-\text{IoU}$ & 0.65 & 0.57 & \textbf{0.55} & 0.56 & 0.56 \\
Center error (\% diag) & 0.09 & 0.04  & \textbf{0.04} & 0.04 & 0.04  \\
Size error (relative) & 0.37 & \textbf{0.27} & 0.31 & 0.32 & 0.32 \\
Image AR error & 0.20 & 0.18 & 0.18 & 0.15  & \textbf{0.15} \\
\addlinespace[2pt]
\multicolumn{6}{c}{\textbf{Content} (micro; higher is better)} \\
\addlinespace[2pt]
Text Content F1 & 0.63 (0.69) & 0.69 (0.73) & \textbf{0.78} (0.78) & 0.76 (0.76) & 0.76 (0.76) \\
\addlinespace[2pt]
\multicolumn{6}{c}{\textbf{Style} (micro; higher is better for style F1 and font accuracies; lower is better for color shifts)} \\
\addlinespace[2pt]
Any-style F1 & 0.55 (0.67) & \textbf{0.77} (0.89) & 0.74 (0.74) & 0.73 (0.73) & 0.73 (0.73) \\
Font Family Acc (micro) & 0.17 (0.27) & 0.33 (0.45) & 0.32 (0.32) & 0.41 (0.41) & \textbf{0.42} (0.42) \\
Font Group Acc (micro) & 0.59 (0.95) & 0.72 (0.98) & 0.98 (0.98) & 0.98 (0.98) & \textbf{0.98} (0.98) \\
Font size MAE (pt) & \textbf{5.93} & 10.18 & 6.22 & 8.92 & 8.97 \\
Text color $\Delta E_{00}$ & 2.27 & 3.37  & \textbf{1.46} & 2.57 & 2.55 \\
Contrast $|\Delta|$ shift & 0.63 & 2.47  & \textbf{0.26} & 0.75 & 0.77 \\
\bottomrule
\end{tabular}}
\caption{\small{Extraction accuracy and geometry quality by model. Each cell shows \emph{end-to-end} and (parsed-only) values, when applicable. Higher is better for F1/accuracy; lower is better for error metrics. Best model metric is boldfaced.}}
\vspace{-1.5em}
\label{tab:extraction-summary}
\end{table*}

\subsection{Slide Deck Narrative Order Performance}\label{app:ordering}

\begin{figure*}[!h] 
  \vspace{-\baselineskip} 
  \centering
  \includegraphics[width=\linewidth]{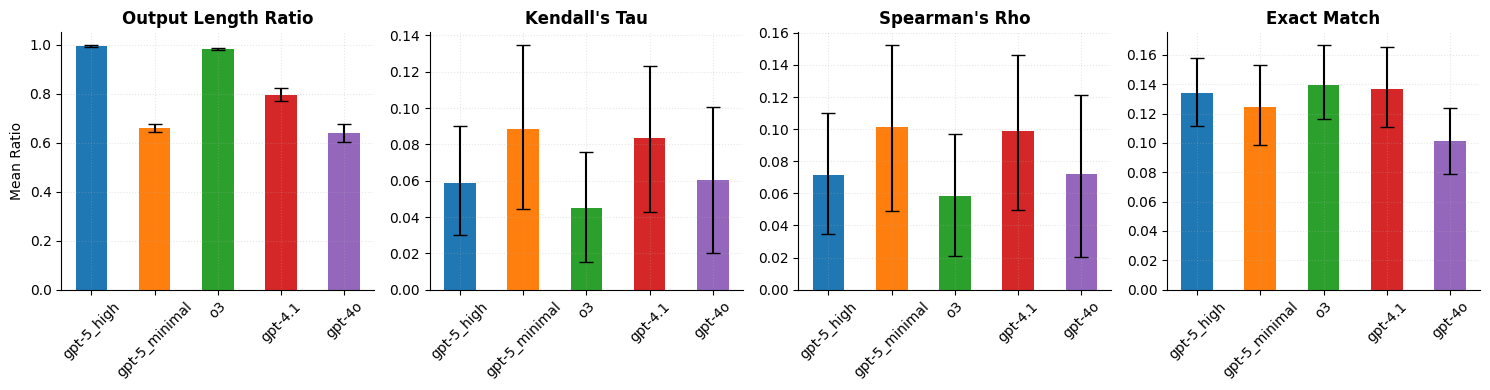}
  \vspace{-2em}
  \caption{\small\textbf{Slide Deck Ordering Prediction}: 1) \underline{Output Length Ratio}: {GPT-5-high} and {o3} successfully generate nearly complete sequences 2) \underline{Kendall's $\tau$} and 3) \underline{Spearman's $\rho$}: despite overlapping confidence integrals, {GPT-5-minimal} and {GPT-4.1} show a consistent upward trend among these two measure, indicating potential robustness that warrants further investigation 4) \underline{Exact Match}: models exhibit similar performance around 0.14 with {GPT-4o} being the lowest.}
  \label{fig:ordering}
\end{figure*}

To assess narrative comprehension, we examine how effectively the VLM reconstructs the original sequence of slides from a randomly shuffled deck (Figure \ref{fig:ordering}). Each deck is segmented into individual slide representations, which are then randomly reordered and input into the model along with a prompt instructing it to restore the correct order. The model’s predicted sequence is evaluated against the ground truth using Kendall’s $\tau$, Spearman’s $\rho$, and normalized exact match metrics. We report the mean and standard deviation across all decks.

As a preliminary step, we verify whether the models can generate output sequences that match the full length of the original presentations. For instance, if a presentation contains 23 slides, the model should produce an ordered list of 23 elements. According to Figure 6 (left), \texttt{GPT-5 high} and \texttt{o3} successfully generate nearly complete sequences, whereas other models struggle to even identify the correct number of slides present in the input.

Focusing on presentations with correctly predicted lengths, {GPT-5-minimal} and {GPT-4.1} demonstrate relatively strong performance in ordering accuracy, as measured by Kendall's $\tau$ and Spearman's $\rho$, particularly outperforming \texttt{o3}.
However, across the board, all models exhibit limited capability in narrative ordering, with scores below 0.15.
This indicates substantial room for improvement before approaching the theoretical upper bound of 1.0 across all metrics.
While the models appear capable of interpreting slide content and multimodal layout, they still face significant challenges in reasoning through the narrative structure.

\section{Fonts and Font Groups Used in the Analysis}
\subsection{Canonicalized Font Names and Counts in the Dataset} 
Table~\ref{tab:font_family_count_data} shows the count statistics of different fonts in text elements present in the ground truth slides.

\begin{table*}[h!]
\centering
\begin{tabular}{lr|lr|lr}
\hline
\textbf{Font} & \textbf{Count} & \textbf{Font} & \textbf{Count} & \textbf{Font} & \textbf{Count} \\
\hline
calibri & 2183 & arial & 1692 & unknown & 460 \\
lato & 260 & montserrat & 203 & roboto & 159 \\
open sans & 132 & century gothic & 105 & oswald & 105 \\
helvetica neue & 98 & avenir next & 97 & garamond & 70 \\
verdana & 66 & ibm plex sans & 65 & corbel & 64 \\
georgia & 61 & source sans pro & 53 & libre franklin & 43 \\
tahoma & 41 & patrick hand & 33 & raleway & 32 \\
soehne & 31 & dosis & 30 & inter & 22 \\
times new roman & 22 & quattrocento sans & 20 & titillium web & 20 \\
bahnschrift & 16 & barlow & 16 & cambria & 16 \\
elephant & 15 & franklin gothic & 14 & nunito & 14 \\
gill sans & 12 & amatic sc & 10 & american typewriter & 10 \\
source code pro & 10 & ubuntu & 9 & ibm plex mono & 5 \\
palatino linotype & 4 & aptos & 3 & handwriting & 3 \\
segoe script & 3 & bookman old style & 2 & menlo & 2 \\
playfair display & 2 & tenorite & 2 & bodoni & 1 \\
inconsolata & 1 & pacifico & 1 & proxima nova & 1 \\
segoe ui & 1 & Total & 6340 &  &  \\
\hline
\end{tabular}
\caption{\small{Frequency of different font families in the ground truth data (sorted descending, row-major)}}
\label{tab:font_family_count_data}
\end{table*}

\subsection{Font $\rightarrow$ Font Group Mapping} \label{appendix:font_group_mapping}
\begin{lstlisting}[style=code]
# Sans
"arial":"sans","calibri":"sans","helvetica":"sans","helvetica neue":"sans","segoe ui":"sans","verdana":"sans",
"tahoma":"sans","gill sans":"sans","inter":"sans","roboto":"sans","open sans":"sans","lato":"sans",
"montserrat":"sans","source sans pro":"sans","libre franklin":"sans","quattrocento sans":"sans",
"ubuntu":"sans","barlow":"sans","bahnschrift":"sans","ibm plex sans":"sans","soehne":"sans","dosis":"sans",
"poppins":"sans","raleway":"sans","titillium web":"sans","nunito":"sans","corbel":"sans","candara":"sans",
"century gothic":"sans","avenir":"sans","avenir next":"sans","franklin gothic":"sans","arial rounded mt":"sans",
# Serif
"times new roman":"serif","georgia":"serif","garamond":"serif","cambria":"serif","palatino linotype":"serif",
"bookman old style":"serif","elephant":"serif","merriweather":"serif","playfair display":"serif",
"bodoni":"serif","bodoni mt":"serif","didot":"serif","tinos":"serif","cmr10":"serif","american typewriter":"serif",
# Mono
"courier new":"mono","courier":"mono","consolas":"mono","menlo":"mono","monaco":"mono","inconsolata":"mono",
"fira mono":"mono","source code pro":"mono","roboto mono":"mono","ibm plex mono":"mono",
# Script / Hand / Display
"comic sans ms":"script","brush script mt":"script","brush script":"script","amatic sc":"script",
"patrick hand":"script","architects daughter":"script","caveat":"script","pacifico":"script","lobster":"script",
"impact":"display","bebas":"display",
# Others
"roboto slab":"serif","carlito":"sans","asana":"serif","tenorite":"sans","aptos":"sans",
"segoe ui emoji":"sans","segoe ui symbol":"sans",
\end{lstlisting}

\subsection{Font Group Frequencies}
Table~\ref{tab:font_group_count_data} shows the count statistics of different fonts in text elements present in the ground truth slides.
\begin{table}[h!]
\centering
\vspace{-1em}
\begin{tabular}{lr|lr|lr}
\hline
\textbf{Font} & \textbf{Count} & \textbf{Font} & \textbf{Count} & \textbf{Font} & \textbf{Count} \\
\hline
sans & 5503 & other & 569 & serif & 203 \\
script & 47 & mono & 18 & Total & 6340\\
\hline
\end{tabular}
\caption{\small{Frequency of different font groups in the ground truth data (sorted descending, row-major)}}
\label{tab:font_group_count_data}
\end{table}

\section{Reproducibility and Safety Checks for Slide Perturbation}
\begin{itemize}\itemsep2pt
\item \textbf{Seeding:} All RNG draws use a fixed base seed; per-slide streams can be derived via a deterministic hash of the slide ID.
\item \textbf{Validity:} Geometry is clamped to the canvas (unless explicitly allowed); sizes are lower-bounded by 1\,px.
Colors are validated to normalized hex (\#RRGGBB) before export.
\item \textbf{No-op at $s = 0$:} We return an unchanged copy when \(s \leq 10^{-12}\).
\item \textbf{On Monotonicity:} Because operations are stochastic, a single draw at $s = 1.0$ need not strictly dominate a draw at $s < 1$, but it does so at expectation (all scales/probabilities are monotone in \(s\)).
\end{itemize}

\section{Declaration of LLM Usage}
We used large language model (LLM) assistants solely for \emph{writing and tooling support}, including (i) manuscript/LaTeX editing, phrasing, and formatting, and (ii) non-substantive code assistance in VS Code (\eg~refactoring, bug fixing, style cleanups, and commenting).
All algorithms, evaluation designs, datasets, metrics, and reported results were specified by the authors; LLM-suggested text/code was reviewed, verified, and tested by the authors before inclusion.
This usage does not impact the core methodology or conclusions.

